\documentclass[10pt,twocolumn,letterpaper]{article}

\usepackage{cvpr}
\usepackage{times}
\usepackage{epsfig}
\usepackage{graphicx}
\usepackage{grffile}
\usepackage{amsmath}
\usepackage{amssymb}
\usepackage{overpic}
\usepackage{amsmath}
\usepackage{amsfonts}
\usepackage{nonfloat}
\usepackage{booktabs}
\usepackage{multirow}

\usepackage{url}
\usepackage{}
\usepackage[pagebackref=true,breaklinks=true,letterpaper=true,colorlinks,bookmarks=false]{hyperref}
\usepackage{textcomp} % for textonehalf
\usepackage{subfig}
\usepackage[font=small,labelfont=bf]{caption}

\usepackage{color, colortbl}

\usepackage{wrapfig}
\graphicspath{{imgs/}{data/renders_turn_table/}{imgs/rendered_results/}{imgs/tabletop_results/}{imgs/nyu_results/}}

\usepackage{xspace}
\renewcommand*{\eg}{e.g.\@\xspace}
\renewcommand*{\ie}{i.e.\@\xspace}
\newcommand*{\ea}{et al.\@\xspace}

\DeclareMathOperator*{\argmin}{arg\,min}

\cvprfinalcopy % *** Uncomment this line for the final submission

\def\cvprPaperID{2604} % *** Enter the CVPR Paper ID here

\newcommand{\mnistocc}{MNIST$_{\text{OCC}}$\@\xspace}

\definecolor{red}{rgb}{0.95,0.4,0.4}
\definecolor{blue}{rgb}{0.4,0.4,0.95}
\definecolor{darkblue}{rgb}{0,0,0.8}
\definecolor{darkred}{rgb}{0.8,0,0}
\definecolor{lightred}{rgb}{1.0, 0.7, 0.7}
\definecolor{darkgreen}{rgb}{0,0.5,0}
\definecolor{grey}{rgb}{0.6,0.6,0.6}

\newcommand{\x}{\ensuremath{\mathbf{x}}\xspace}
\newcommand{\y}{\ensuremath{\mathbf{y}}\xspace}
\newcommand{\Y}{\ensuremath{\mathcal{Y}}\xspace}
\newcommand{\cc}{\ensuremath{\mathbf{c}}\xspace}
\newcommand{\C}{\ensuremath{\mathcal{C}}\xspace}

\renewcommand{\paragraph}{\vspace{2pt}\noindent\textbf}
\usepackage[hang,flushmargin]{footmisc}

\title{DiverseNet: When One Right Answer is not Enough$^\dagger$}

\author{
Michael Firman\,\textsuperscript{1*} \qquad 
Neill D.\,F. Campbell\,\textsuperscript{2} \qquad 
Lourdes Agapito\,\textsuperscript{1} \qquad 
Gabriel J. Brostow\,\textsuperscript{1*}  \\ 
\textsuperscript{1}
\emph{University College London} \qquad
\textsuperscript{2}
\emph{University of Bath} } % \\ {\url{http://visual.cs.ucl.ac.uk/pubs/DiverseNet}}}
\begin{document}

\maketitle
\let\thefootnote\relax\footnotetext{\noindent $^{*}$MF is now with Niantic, Inc. GB is now with UCL and Niantic, Inc. \\$^\dagger$ This paper was published at CVPR 2018.}

\begin{abstract}
\vspace{-3pt}
Many structured prediction tasks in machine vision have a collection of acceptable answers, instead of one definitive ground truth answer. Segmentation of images, for example, is subject to human labeling bias. Similarly, there are multiple possible pixel values that could plausibly complete occluded image regions. State-of-the art supervised learning methods are typically optimized to make a single test-time prediction for each query, failing to find other modes in the output space. Existing methods that allow for sampling often sacrifice speed or accuracy.

We introduce a simple method for training a neural network, which enables diverse structured predictions to be made for each test-time query. For a single input, we learn to predict a range of possible answers. We compare favorably to methods that seek diversity through an ensemble of networks. Such stochastic multiple choice learning faces mode collapse, where one or more ensemble members fail to receive any training signal.
%In this work, we propose several solutions to the mode collapse problem.
Our best performing solution can be deployed for various tasks, and just involves small modifications to the existing single-mode architecture, loss function, and training regime.
We demonstrate that our method results in quantitative improvements across three challenging tasks: $2$D image completion, $3$D volume estimation, and flow prediction.

% We introduce a simple method for training a neural network that enables diverse structured predictions to be made for each test-time query. Where multiple ground truth labels are present at training time for a single item, we can exploit these to better learn to predict a range of possible answers. Our method is straightforward and is easily added to existing neural network architectures with only a few extra parameters to learn. Across a range of tasks, including image completion, colorization, segmentation, and flow estimation, we demonstrate that DiverseNet produces a variety of good predictions, including ones associated with a minority of the training data. %We evaluate our method on a diverse range of tasks including image completion, colorization, segmentation, and flow estimation.
\end{abstract}

\vspace{-8pt}

%%%%%%%%%%%%%%%%%%%%%%%%%%%%%%%%%%%%%%%%%%%%%%%%%%%%%%%%%%%%%%
% \note{Consider citing  Photographic Image Synthesis with Cascaded Refinement Networks}

\section{Introduction}\label{Sec:Intro}

\begin{figure}[!t]
%\begin{flushleft}
\centering
% \noindent
\captionsetup[subfigure]{width=\columnwidth}
\subfloat[Many prediction tasks have ambiguous interpretations. For example, giving a 2D rendering to a human, there are different possible 3D interpretations.]{\includegraphics[width=0.85\columnwidth]{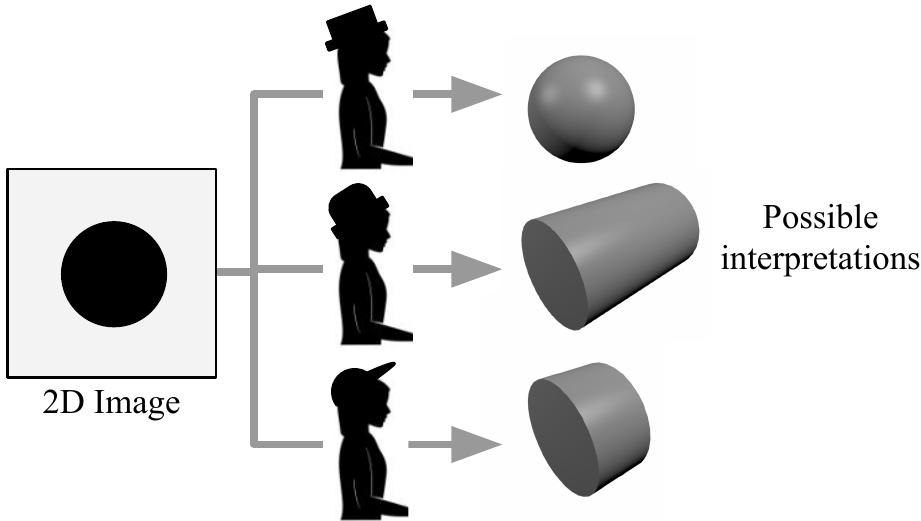}}\\[5pt]
\subfloat[Standard CNNs trained under multi-modal labels tend to blur together or ignore the distinct modes. We introduce a modification to standard neural networks that gives the user a control parameter $\mathbf{c}$.
Different values of $\mathbf{c}$ produce diverse structured outputs for the same input.]{\includegraphics[width=0.8\columnwidth]{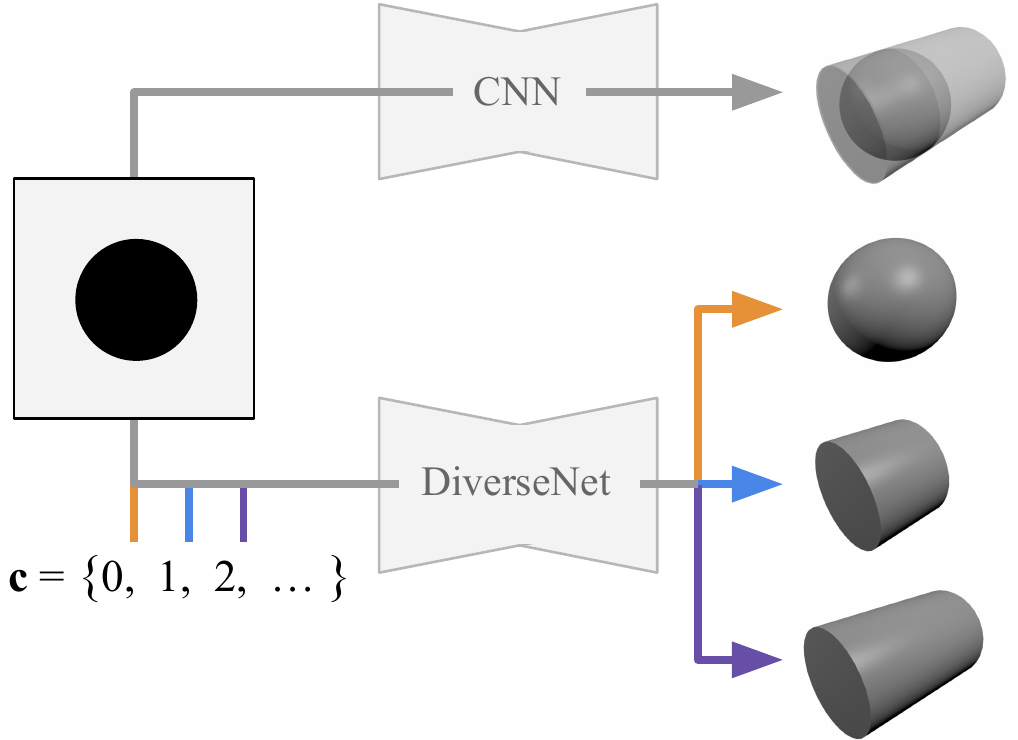}} \\[-2pt]

\caption{\label{fig:banner} Our easy modification enhances typical loss functions, producing networks that predict multiple good answers, not just one best-majority-fit output. Access to ``minority'' outputs is especially useful in applications dealing with ambiguous completions or human preferences.}
\vspace{-15pt}
\end{figure}

Computer vision systems are typically trained to make a single output prediction from a given input.
However, in many cases, there is more than one correct answer.
Consider the case of $3$D projection in Figure~\ref{fig:banner}. 
We wish to infer what $3$D shape projected to form a given $2$D silhouette. 
The training data suggests that there should be more than one answer: some users identify the circle as being produced by a sphere, while others hypothesized different sized cylinders, viewed head-on.
We assert that \emph{each} of these interpretations is correct, and, depending on the application scenario, different outputs may be required.
However, a typical neural network predictor might make a single prediction which averages together the modes present (Figure~\ref{fig:banner}(b), top).

While many machine learning systems are able to produce samples at test time (\eg \cite{gal-arxiv-2015, kingma-iclr-2014, salakhutdinov-jmlr-2009}), we propose a method which explicitly exploits diversity in \emph{training labels} to learn how to make a range of possible structured predictions at test time.
Our system is an add-on modification of the loss function and training schedule that is compatible with any supervised learning network architecture. % of any neural network trained for supervised learning.
Our contributions allow for a network to take as input a test image \x and a \emph{control parameter} \cc.
Where training time diversity exists, our loss encourages the network to find the different modes in the label space.
At test time, providing the same \x with different values of \cc produces different predictions (Figure~\ref{fig:banner}(b)).
Our method can be applied to any supervised learning architecture and loss.

Our method is also applicable in cases where there \emph{is} one definitive ground truth answer.
For example, a grayscale image has a single ground truth colorization; however, for most applications a user may be satisfied with a range of plausible suggestions \cite{luan-eg-2007, marks-acm-1997}.
Our method can predict diverse solutions even where only one label exists for each training item. 

Our main contribution is an architecture-modification and loss which prevent mode-dropping. Mode-dropping is a phenomenon recently observed in GAN training~\cite{srivastava-arxiv-2017}, where regions of output space are not reached by predictions from a model.
We observe this effect in \cite{lee-nips-2016}; during training, their method can result in some members of the ensemble failing to receive any training signal. 
This occurs when other ensemble members are much closer to the mean of the output space. 
At test time, those ensemble members fail to make a meaningful prediction. Our method avoids this problem.

%%%%%%%%%%%%%%%%%%%%%%%%%%%%%%%%%%%%%%%%%%%%%%%%
\section{Related work}

There is a large body of work that examines the cases where labels for data are diverse, noisy or wrong~\cite{krause-eccv-2016, misra-cvpr-2016, sharmanska-cvpr-2016}.
% Multiple labels for a single data point are often aggregated into one label, sometimes taken as the mode or average of the noisy labels.
% Some works,  \eg \cite{sharmanska-cvpr-2016}, instead scale the influence of each label according to label consensus when training a classifier.
Most of these, however, assume that the labels are a noisy approximation of one true label, while we assume they are \emph{all} correct.
Methods which make diverse predictions can be roughly categorized as:
(a) those which allow for \emph{sampling} of solutions;
(b) \emph{ensembles} of models, each of which can give a different prediction,
and
(c) systems which find diverse predictions through \emph{test-time} optimization.

\paragraph{Sampling methods}
Where parameters are learned as parametric distributions, samples can be drawn; for example, consider the stochastic binary distribution in the case of Boltzmann Machines~\cite{Ackley-BM}.
Restricted Boltzmann Machines~\cite{Hinton-RBM}, Deep Boltzmann Machines~\cite{salakhutdinov-jmlr-2009} and Deep Belief Networks (DBNs)~\cite{Hinton-DBN} are all generative methods that learn probabilistic distributions over interactions between observed and hidden variables. 
Such probabilistic networks allow samples to be drawn from the network at test time using MCMC methods such as Gibbs Sampling. 
This has been used, for example, to sample 2D and 3D shape completions \cite{Eslami-ShapeBM,wu-cvpr-2015}. Unfortunately, these sampling processes are often time consuming resulting in models that are difficult to train (as the models scale) and expensive to sample.

Removing the stochastic nature of the units, the supervised learning scheme for DBNs leads to autoencoders~\cite{Hinton-AE} that are easier to train but the directed model no longer maintains distributions and therefore cannot be sampled from.  
% allow for a Gaussian distribution to be learned
Variational autoencoders (VAEs)~\cite{kingma-iclr-2014} use a variational approximation to estimate (Gaussian) distributions over a low dimensional latent space as a layer in the predictive network. These distributions capture uncertainty in the latent space and can, therefore, be sampled at test time. However, the smooth local structure of the latent space makes it unlikely to capture different modes; instead the variational approximation is targeted towards complexity control on the dimensionality of the latent space.
%VAEs generally restrict the choice of architecture --- their sampling step can't be used with fully convolutional networks, or incorporate skip connections between the encoder and decoder. \note{need to verify this claim}.

Generative adversarial networks (GANs) have an optimization scheme which enables novel data to be sampled ~\cite{goodfellow-arxiv-2014}, possibly conditioned on an input sample~\cite{mirza-arxiv-2014}.
%Reed \ea \cite{reed-icml-2016} use this to generate multiple samples of 2D images conditioned on an input sentence and a noise vector.
%While walks through the $z$ space gives interpolation between predictions, no single dimension in $z$ gave explicit control.
\emph{Unsupervised} control can be enabled with an extra input, which is encouraged to correlate with the generated image~\cite{chen-arxiv-2016}.
This method is restricted to use with GANs, while our method can be added to any supervised loss.
Like~\cite{chen-arxiv-2016}, though, we find the relationship between the controlling input and modes in the output space automatically, and we control the output space via an additional input~(\cc).

%In spite of suggestions such as \cite{salimans-arxiv=2016}, generative adversarial networks are considered notoriously difficult to optimize \cite{pfau-arxiv-2016}.
%and are generally difficult to use on large images.
%, have not been extended to creating very large images, and have not been demonstrated on problems beyond natural image generation.
%In contrast, our `meta'-loss function is a simple modification of any standard loss, and we demonstrate its applicability to a range of challenging tasks.

\begin{figure*}[!t]
\centering
  \includegraphics[width=1.0\textwidth]{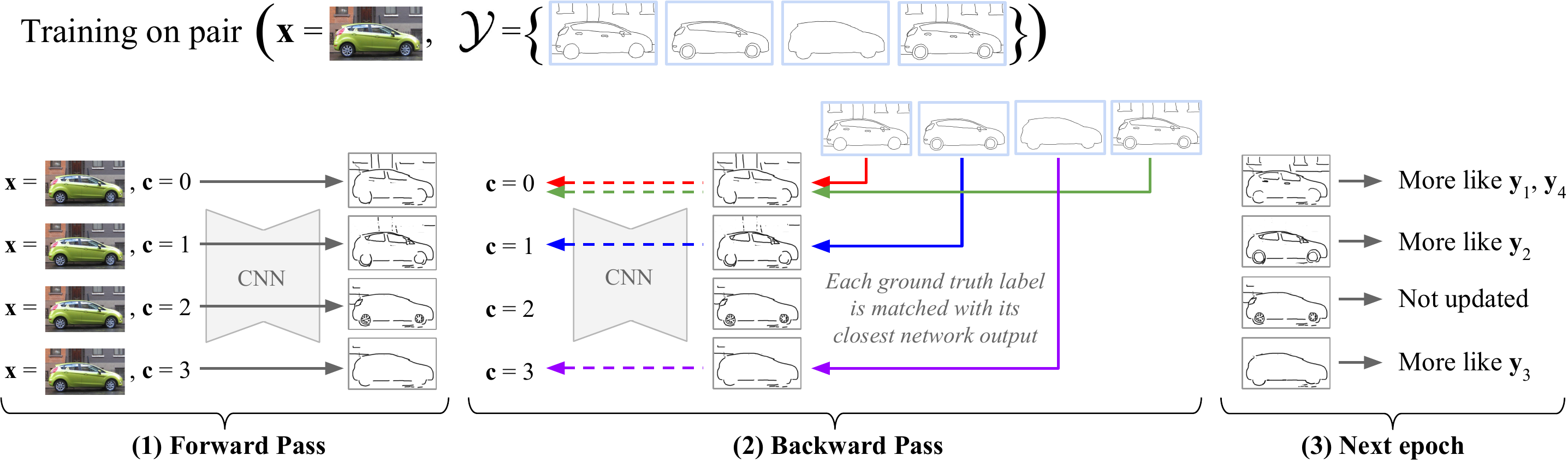}%
  \vspace{-5pt}
  \caption{\small  A toy example showing our update formulation as applied to image segmentation.
  Here, the image \x has four \y labels associated with it.
  \textbf{(1) Forward pass} On the forward pass, the same \x is provided to the network $N$ times, each time with a different value for \cc.
  Each of the outputs produced is different due to the different values of \cc.
   \textbf{(2) Backward pass}
   We associate each ground truth label $\y \in \Y$ with its best matching network output. This matching is indicated by the solid arrows. Losses are then backpropagated only for these values of \cc. The paths along which gradient flows are shown by dashed arrows.
   \textbf{(3) Next epoch}
   When the same \x and \cc values are passed to the network on the next epoch, each prediction is now a little more like the ground truth image it was matched with.}
  \label{fig:example_update}
  \vspace{-7pt}
\end{figure*}

\paragraph{Ensembles}
Ensembles of neural networks have been found to outperform networks in isolation~\cite{krizhevsky-nips-2012}.
%\remove{For example Krizhevsky \ea \cite{krizhevsky-nips-2012} reduced the top-1 error rate on Imagenet by 2.6\% using an ensemble of 5 CNNs.}
Typically, each network is trained on all the training data, allowing randomness in initialization and augmentation to lead each network to a distinct local minimum.
Bagging, where each classifier is trained on a random subset of the training data, can be used to increase the variance of prediction from multiple weak classifiers~\cite{breiman-ml-2001}.
%One way to increase the variance of predictions from multiple weak classifiers can be increased with bagging, where each classifier is trained on a random subset of the training data \cite{breiman-ml-2001}.
Alternatively, Liu and Yao~\cite{liu-smc-1999} \emph{explicitly} encourage diversity in the ensemble by the addition of a variance term to the loss function that forces solutions apart from each other, and similarly Dey~\ea~\cite{dey-iccv-2015} train a sequence of predictors explicitly to give diverse predictions.
As far as Dropout~\cite{srivastava-jmlr-2014} can be considered to approximate an ensemble~\cite{baldi-nips-2013}, then its application at \emph{test time} \cite{gal-arxiv-2015} can be considered as drawing samples from a large ensemble.

%Guzman-Rivera~\ea~\cite{guzman-nips-2012} find diverse solutions as multiple outputs from a structured support vector machine, trained with a \emph{multiple-choice learning} (MCL) loss.
Lee~\ea~\cite{lee-arxiv-2015, lee-nips-2016} introduce a loss which encourages ensemble diversity.
They backpropogate the loss for each training example only through the ensemble member that achieves the best loss on the forward pass.
Each network, over time, becomes an ``expert'' in a different field.
Their loss, which is based on \cite{guzman-nips-2012}, is related to ours.
We differ in three ways.
First, they train an ensemble (or quasi-ensemble, `treenet')  of networks, while we make multiple predictions from a single network, significantly reducing the number of parameters.
Second, their loss does not directly take advantage of training time diversity, while ours handles cases both where we do and do not have multiple labels for each training item. Third, as introduced in Section~\ref{Sec:Intro}, our approach prevents mode-dropping \cite{srivastava-arxiv-2017}.
%, where some modes in output spacae, a limitation observed in GAN training~\cite{srivastava-arxiv-2017}.

% Our main contribution, though, is an architecture and loss which prevents mode-dropping.
% Mode-dropping is a phenomenon recently observed in GAN training \cite{srivastava-arxiv-2017}, where regions of output space are not produced by predictions from a model.
% We observe this effect in \cite{lee-nips-2016}; during training, their method can result in some members of the ensemble failing to receive any training signal. 
% This occurs when other ensemble members are much closer to the mean of the output space.
% At test time, these ensemble members fail to make a meaningful prediction.

%Also, while samples from their ensemble must be discrete (it is impossible to sample partially from one network and partially from another) our method is general enough to allow control with a continuous variable.

\paragraph{Test-time diversity}
Some methods explicitly enforce diverse modes at \emph{test time}.
For example, Batra~\ea~\cite{batra-eccv-2012} find diverse solutions for a Markov random field with a greedy method which applies a penalty to each new solution if it agrees too much with previously discovered modes.
This was subsequently developed to a more general solution \cite{kirillov-iccv-2015}.
In contrast, ours learns at \emph{training time} how to make diverse predictions; each prediction is then made with a single network evaluation.

%These often work via optimization, finding solutions which do not correlate with each other.
%\remove{Similarly, multiple \emph{geometrical} models explaining a scene can be found by encouraging each model to explain a separate set of data points \cite{isack-ijcv-2012}.}

%Where solutions are found with a sampling scheme such as MCMC (\eg \cite{rezende-arxiv-2016} for volume completion), different local minima can reveal different solutions.
%However, these methods can be computationally expensive for high-dimensional structured outputs.

%Nonparametric retrieval allows for multiple options to be retrieved, \eg via hash lookup \cite{lee-acm-2011} or nearest neighbors.
%Non-maximal suppression is often used to get diverse object proposals in detection systems \cite{chen-nips-2015}.

%Zhu~\ea~\cite{zhu-acm-2014} use a predictive method to find modes (clusterings) in a large set of images, for visual exploration.
%also `generating multiple hypotheses for human 3d pose'}
% \todo{Make clear which are our baselines}

%%%%%%%%%%%%%%%%%%%%%%%%%%%%%%%%%%%%%%%%%%%%%%%%
\section{Method}

Typically, training data for supervised learning consists of a set of pairs $\mathcal{D} = \{(\x, \y)\}$.
Each pair consists of an input \x, which in vision tasks is often an image, and the desired output label \y; for our structured prediction tasks, \y is multidimensional.
Training the parameters of a machine learning system $f$ then involves minimizing a loss $l$ summed over this set of data, \ie
\begin{equation}
L = \sum_{(\x, \y) \in \mathcal{D}} l(f(\x), \y).
\label{eqn:standard_loss}
\end{equation}
In our work, we assume that for each \x there is a \emph{set} of labels.
Each training pair $(\x, \Y)$ now comprises a single \x with a set of target values $\Y = \{\y_1, \y_2, \ldots \y_N \}$.
For example, for image segmentation each image may have had boundaries drawn by multiple human labelers.
Note that $N$ can be different for different training pairs.
A straightforward modification of (\ref{eqn:standard_loss}) to minimize the loss over $\Y$ is
\begin{equation}
L = \sum_{(\x, \Y) \in \mathcal{D}} 
\,\,
\sum_{\y \in \Y}
l(f(\x), \y).
\label{eqn:stupid_way}
\end{equation}
Unfortunately, using (\ref{eqn:stupid_way}) explicitly (or averaging the label set, which is often done in practice), results in predictions being made that lie between modes in the label space.
This ``mode collapse'' effect has been observed when training networks on ambiguous tasks like image completion \cite{pathak-cvpr-2016}.

Our model instead accepts as input \x together with a control variable $\cc \in \C$.
Specifically, \cc is fed to the network through concatenation with activations from both dense and convolutional layers, and integer values for $\cc$ are first converted to a one-hot representation.
At test time, users can create different outputs for the same $\x$ by varying the value of \cc (Figure~\ref{fig:banner}(b)).
%Where applicable, this is done after batch normalization and activations.

We assume that, during training, each $\y \in \Y$ is a valid output that we wish our system to be able to reconstruct under at least one value of \cc.
For a single $\x, \y$ pair, the value of \cc which produces a network output that most closely matches \y is 
$\argmin_{\cc \in \C} \,\, l(f(\cc, \x), \y)$.
We want each $\y \in \Y$ to be well reconstructed, so we penalize any $\y$ which is not well reconstructed by $f$.
Our loss is therefore
% \vspace{-5pt}
\begin{equation}
L_\text{div} = 
\sum_{(\x, \Y) \in \mathcal{D}}
\,\, \sum_{\y \in \Y}
\,\, \min_{\cc \in \C} 
\,\, l(f(\cc, \x), \y).
\label{eqn:our_loss}
\end{equation}%
% \vspace{-10pt}
% \huge\Y

We show an illustrative example of an update for a single $(\x, \Y)$ in Figure \ref{fig:example_update}.
%Note that in reality, many such pairs are used to perform a single minibatch update.
On the forward pass through the network, the same \x is passed with different values of \cc.
At this stage in the training, the network produces different results for each value of \cc.
We then associate each possible label, \ie each $\y \in \Y$ with just one of the network outputs.
The losses for these ground truth labels are then backpropagated through the network to the appropriate value of \cc.
After the update, each value of \cc is more closely associated with a different mode in the data.
This update formulation is a multi-label, single-network extension of \cite{lee-arxiv-2015, lee-nips-2016}.

Equation \ref{eqn:our_loss} can be implemented easily.
For each group of inputs to the network, we compute a matrix measuring the pairwise loss between each network output and each $\y \in \mathcal{Y}$.
The \texttt{min} down each column gives the closest matching prediction to each \y.
The sum over all such \texttt{min} values gives the objective $L_\text{div}$.
%, and ensures that the loss is backpropogated to the appropriate examples in the minibatch.

\vspace{5pt}
\paragraph{Preventing mode collapse}
We sometimes find that when $|\C |$ is larger than the number of modes naturally present in the data, then one or more values of $\cc$ can produce poor solutions.
This occurs when, during stochastic training, the modes in the label space are better captured by a subset of \cc values, and therefore other values for \cc are never encouraged to produce any meaningful result.
%When this happens early on in training, this can cause the network to fail to produce meaningful predictions with many \cc values.
To remove this undesirable effect, and to help ensure that no predictions are degenerate, we propose an additional term which ensures that the \emph{worst performing prediction} gets updated, regardless of its proximity to any ground truth label, as
\begin{equation}
L_\text{catchup} = 
\frac{1}{|\C|}
\sum_{(\x, \Y) \in \mathcal{D}}
\,\, \max_{\cc \in \C} 
\,\, \min_{\y \in \Y}
\,\, l(f(\cc, \x), \y).
\label{eqn:extra_loss}
\end{equation}
The $\min$ in (\ref{eqn:extra_loss}) finds, for each network prediction, the distance to its closest matching ground truth label in $\Y$.
The $\max$ then ensures that only the value of $\cc$ corresponding to the \emph{largest} of these losses is updated.
%We explore the impact of including $L_\text{catchup}$ in the experiments section.

\label{sec:beta}
Our final training loss is simply $L = L_\text{div} + \beta L_\text{catchup}$, where $\beta$ is a parameter which trades off the quality of the best reconstructions ($L_\text{div}$) with the worst ($L_\text{catchup}$).
All the results in this paper are produced with $\beta=1$; we examine the effect of tuning this parameter in Section \ref{sec:beta_eval}.
%the loss to the correct values of \cc using a \texttt{min} operation down each column of this matrix.

%\subsection{Diversity without multiple training labels}
% \label{label:div_without_multiple}

%, and because we train a single network we find that each .
%, and the ability for each m %to learn with a continuous value of $\cc$, as we now show.

\vspace{5pt}
\paragraph{Learning without training-time diversity}
Given only a single label for each training image, \ie $| \Y | = 1$, we can still use our system to make diverse predictions.
Our loss function now becomes:
\begin{equation}
L_\text{div} = 
\sum_{(\x, \y) \in \mathcal{D}}
\,\, \min_{\cc \in \C} 
\,\, l(f(\cc, \x), \y).
\end{equation}
This shares similarities with a single-network version of the multiple choice learning loss of \cite{guzman-nips-2012, lee-arxiv-2015}.
However, our formulation shares parameters across each value of \cc, resulting in considerably fewer parameters than an ensemble.

\paragraph{Architecture details}
We insert categorical \cc into the network by converting to one-hot encoding, and concatenating with the activations (Figure~\ref{fig:arch}).
We can also use continuous or multi-dimensional values for \cc.
A continuous \cc could be useful for example when exploring continuous qualitative artistic options, \eg using a deep-learned `Photoshop' to complete missing image regions.
In these cases it becomes intractable to enumerate all possible values in a single minibatch.
Instead, we take a stochastic approximation to (\ref{eqn:our_loss}), and set $\C$ as a set of \emph{samples} from a user-specified distribution.
Our method is applicable to all supervised network architectures, including those with skip connections, dropout, and batch normalization.

%%%%%%%%%%%%%%%%%%%%%%%%%%%%%%%%%%%%%%%%%%%%%%%%%%%%%%%%%%%%%%%%%

\begin{figure}[!b]
% \centering
  \vspace{-6pt}
  \includegraphics[width=1.0\columnwidth]{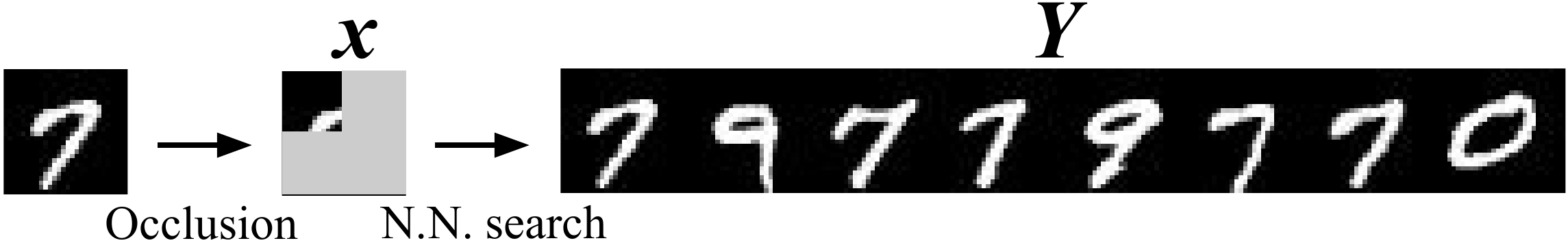}
%   \vspace{-3pt}
  \caption{\small Creating \mnistocc. 
  Each original MNIST digit (left) is occluded to create an \x image.
  The $8$ matching digits which most closely match \x form the set \Y.
  }
  \label{fig:mnist_occ_creation}
\end{figure}

\begin{figure*}[!t]
  \raisebox{100pt}[0pt][0pt]{
  \renewcommand{\arraystretch}{1.65}
  \small  \noindent\hspace{-10pt}\begin{tabular}{  l  }
      \x \emph{(Network input)} \\ 
      \Y \emph{(Ground truth)} \\
    (A) $L_2$  \\
(B) Dropout  \\
(C) Bagged  \\
(D) Lee \ea  \\
(E) Lee \ea + $\epsilon L_2$  \\
(F) GAN sample  \\
(G) Cond. GAN  \\
(H) VAE  \\
Ours \\
  \end{tabular}}\hspace{-5pt}
  \includegraphics[width=4.9cm]{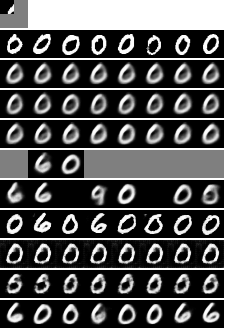}\hspace{-2pt}%
  \includegraphics[width=4.9cm]{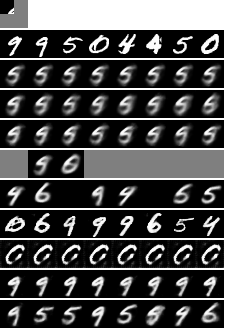}\hspace{-2pt}%
  \includegraphics[width=4.9cm]{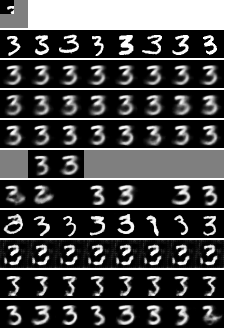}\vspace{-2pt}%
   \vspace{-2pt}
  \caption{\small Predictions from the models.
  For each model, described in Section \ref{sec:mnist_eval}, we make $8$ predictions on each image.
  The ground truth row shows the \Y values in MNIST$_{\text{OCC}}$, found from the test set using nearest neighbor lookup --- the left-most value in \Y is the image from which \x was generated.
  More results are given in the supplementary material. \\
  %*Note that these results were produced from a modified version of the baseline algorithm. See text for details.
  }
  \vspace{-16pt}
  \label{fig:mnist_results}
\end{figure*}

%%%%%%%%%%%%%%%%%%%%%%%%%%%%%%%%%%%%%%%%%%%%%%%%%%%%%%%%%%%%%%%%%
\section{Experiments}

Most networks can be augmented to become a DiverseNet, so we perform experiments across a range of datasets, loss functions, and applications, to demonstrate the generalizability of this approach. 
We validate that it copes with non-unique labelings and improves diversity against competing diversity-seeking methods. 
Further, we establish that our models have fewer parameters (almost as few as a native network) and so are easier to train than methods like~\cite{lee-nips-2016}. 
Nonetheless, we sacrifice very little accuracy compared to single-best models (Note: like class-imbalance problems, some reduction in raw accuracy is expected).

\paragraph{Evaluating diverse predictions}
Where only a single ground truth answer exists, the \textbf{k-best oracle}~\cite{guzman-nips-2012,lee-nips-2016} is a suitable scheme for evaluating diverse predictions.
For each input, $k$ random predictions are made, and the one which most closely matches the ground truth is chosen. This error is averaged over all test inputs to report the overall error for a particular value of $k$.
Sweeping $k$ allows us to plot a graph, where error typically reduces as $k$ is incremented and more predictions are made.

Where there are \emph{multiple} ground truth answers for each data point, a perfect algorithm would generate each of the ground truth answers. 
An error is computed for each possible ground truth answer, where again we compare to the closest match among $k$ predictions from the model.
The final error is the mean of all these errors.

%, we introduce a new \textbf{k-all oracle} evaluation method. % for datasets. 

\subsection{Comparing to baselines on \mnistocc}
\label{sec:mnist_eval}

We use the MNIST dataset \cite{lecun-misc-1998} to demonstrate how our system can be used to find distinct outputs where there is ambiguity at training time.
We create a modified \emph{occluded} version of the dataset, \mnistocc, designed for training and evaluating image completion where there are multiple correct answers (Figure \ref{fig:mnist_occ_creation}).
Each \x in \mnistocc comprises an original MNIST digit, where all pixels are set to zero \emph{except} the $14 \times 14$ square in the top-left corner of the image.
For each \x we must synthesize a diverse set of labels \Y.
After cropping, we find the $8$ closest matching \x values in the corresponding training set. 
Their respective associated \y values form the set \Y.
The aim of this image completion task is to accurately recover \Y given a single \x.

\begin{figure}[b]
\vspace{-10pt}
  \includegraphics[width=0.75\columnwidth]{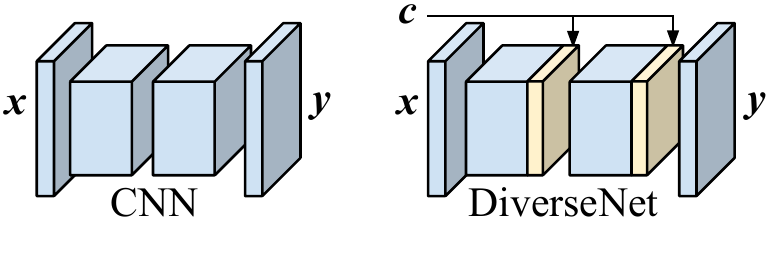}%
  \vspace{-13pt}
  \caption{\small DiverseNet's architecture differs from a standard CNN through concatentations of \cc with the network activations. \cc is typically a one-hot encoding of the integer parameter.}
  \label{fig:arch}
\end{figure}

For all \mnistocc experiments, unless stated otherwise, we use a simple bottleneck architecture (Figure~\ref{fig:arch}).
We train our algorithm with $N=8$ discrete values for \cc, and for all competing methods we draw $8$ samples.
Networks which use \cc as input have it concatenated with the activations as a one-hot encoding, as described above.
For each baseline, we train with \x against all values of \Y.
We compare against the following baselines:

% \begin{table}[!b]
% \footnotesize
% \begin{tabular}{|l|c|c|c|c|}
% \hline
% & \textbf{Params.} & \textbf{$\times$ (A)}
% & \textbf{Test-time (ms)} & \textbf{$\times$ (A)} \\
% \hline
% (A) $L_2$ loss & & 1 & & 1.0 \\
% (B) Dropout & & 1 & & 1.0 \\
% (C) Bagged & & $k$ & & 1.0 \\
% (D) M-heads & & $k$ & & 1.0 \\
% (E) GAN & & 1 & & 1.0 \\
% (F) CondGAN & & 1 & & 1.0 \\
% (G) DBM & & 1 & & 1.0 \\
% (H) VAE & & 1 & & 1.0 \\
% \hline
% \end{tabular}
% \vspace{5pt}
% \caption{\small A comparison of the model size and test-time speed of MNIST image completion alternatives.
% The number of parameters in the network and the average time for a single completion mode are given.
% We also give these numbers as a multiple of the values in the first row.}
% \end{table}

\begin{figure}
  \includegraphics[width=1.0\columnwidth,clip=true,trim={0.5cm 0.5cm 1.65cm 0.5cm}]{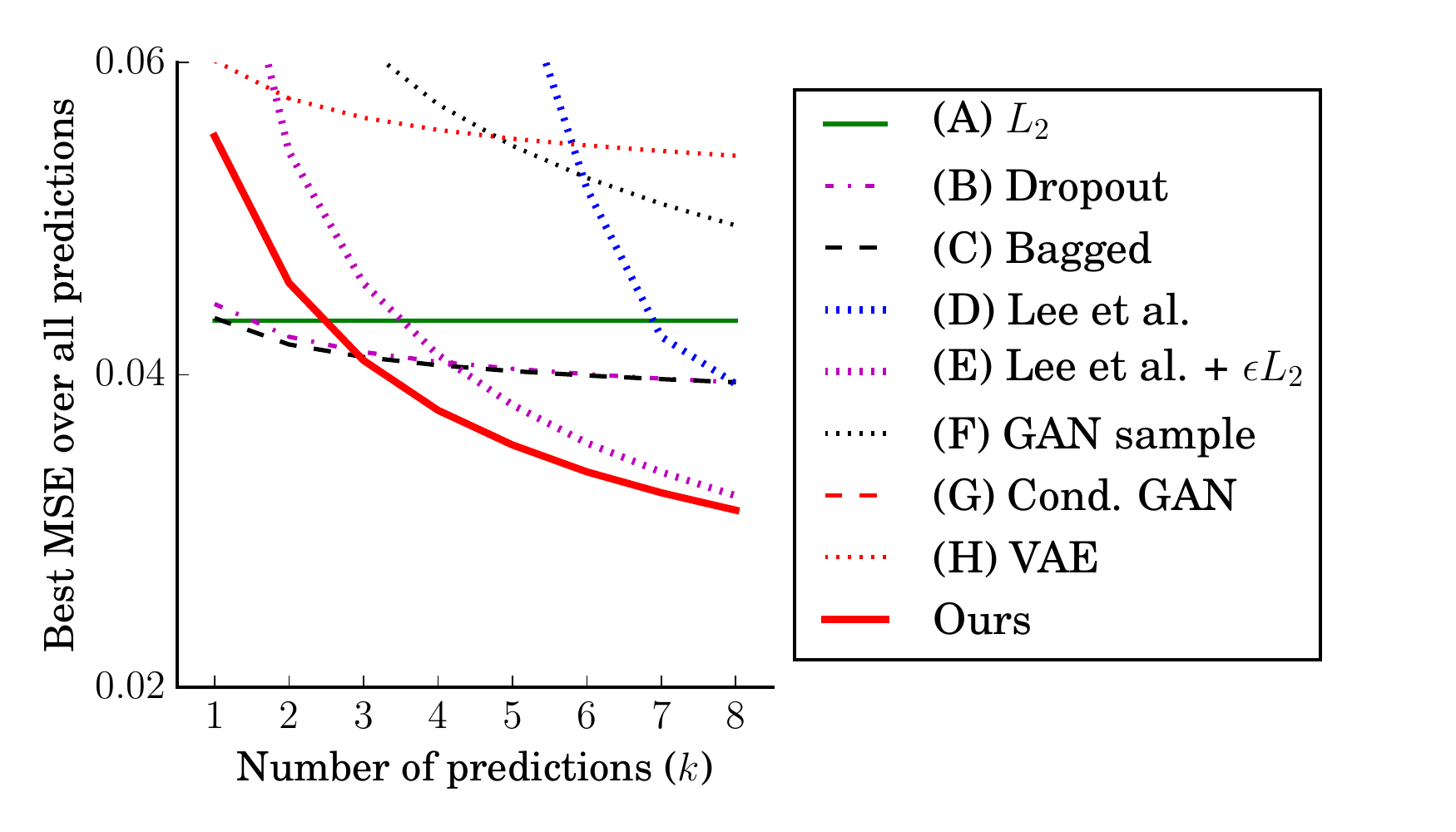}%
  \vspace{-5pt}
  \caption{\small Quantitative results on \mnistocc. 
  As more predictions are made from the model, the oracle performance of all systems except (A) $L_2$ improves.}
  \vspace{-15pt}
  \label{fig:mnist_quant}
\end{figure}

\paragraph{(A) Standard $L_2$ loss}
A bottleneck architecture, trained with $L_2$ reconstruction loss between the input and output.

\paragraph{(B) Test-time Dropout \cite{gal-arxiv-2015,srivastava-jmlr-2014}}
We train a single network with Dropout applied to the dense layers.
At test time, we sample $N$ predictions with different dropout values.

%We also tried Dropout on the convolutional layers, but this gave worse results.

\paragraph{(C) Bagged ensemble}
We train an ensemble of $N$ networks, each trained on a random $2/3$ of the training set.
At test time, each network gives a single prediction.

\paragraph{(D) Lee \ea ensemble}
We form a `treenet' ensemble trained in unison as described in \cite{lee-arxiv-2015} (see Section 2); each ensemble member shares weights in the encoder, but has separate weights in the dense layers and decoder.

\paragraph{(E) Lee \ea ensemble + $\epsilon L_2$}
We found that several ensemble members in (D) failed to learn, giving null predictions.
By adding a small ($10^{-4}$) amount of $L_2$ loss we reduced the number of degenerate members and improved their scores.

%At training time, each training image is passed through all the ensembles.
%The training loss is backpropagated through the network which performs the best.

\paragraph{(F) GAN samples}
%GANs \cite{goodfellow-arxiv-2014} in their vanilla form cannot produce samples conditioned on an input.
For this simple baseline we sample $20,000$ images from a GAN \cite{goodfellow-arxiv-2014} trained on MNIST.
For each test \x, the $k$ samples which most closely match the unmasked corner are taken as the predictions.

\paragraph{(G) Conditional GAN \cite{mirza-arxiv-2014}}
Here the generator network is trained to produce samples conditioned on \x.
We use our bottleneck architecture, with a noise vector concatenated on the bottleneck and batch normalization  \cite{ioffe-corr-2015} for stability.
Different noise samples give each test-time completion.

% \paragraph{Conditional InfoGAN}
% As conditional GAN, but we also include a discrete $c$, which can take on $N$ values.
% Our samples now come from different values of $c$.

\paragraph{(H) Variational Autoencoder}
Here we make the architectural change of including a Gaussian sampling step in the bottleneck, implemented and trained as \cite{kingma-iclr-2014}.
Test-time samples are produced by sampling from the Gaussian.

%\paragraph{(I) Deep Boltzmann Machine} \cite{salakhutdinov-jmlr-2009} \todo{}

\vspace{10pt}

Quantitative results are shown in Figure~\ref{fig:mnist_quant}. When making just one or two predictions, our method has a higher MSE than methods trained with an $L_2$ loss.
However, beyond $3$ predictions, we outperform all competitors.
Figure~\ref{fig:mnist_results} shows qualitative results.
The bagged ensemble (C) performs well, as expected, while (D) \cite{lee-arxiv-2015} produces poor results.
Qualitative inspection shows that, on this task, some of their ensemble members never produce meaningful results.
We see how their method is hurt by dropping modes.
Our $L_\text{catchup}$ loss helps to prevent this mode dropping.
%Both the improved loss function and our single-network design helps avoid degenerate predictions from values of \cc --- we demonstrate this experimentally in Section \cite{sec:shapenet}.
We note that conditional GANs produce highly correlated samples, as observed in \cite{isola-arxiv-2016,mathieu-iclr-2016}, since the network learns to ignore the noise input.
The sharpness of (F) \emph{GAN sample} suggests that an adversarial loss could help improve the visual quality of our results.

\begin{figure}
  \raggedright
  \small  \hspace{2pt} \x \hspace{53pt} \Y \hspace{48pt} $-1 \,\, \longleftarrow \,\, \longleftarrow \,\, \cc \,\, \longrightarrow \longrightarrow \,\, 1$
  \includegraphics[width=1.0\columnwidth]{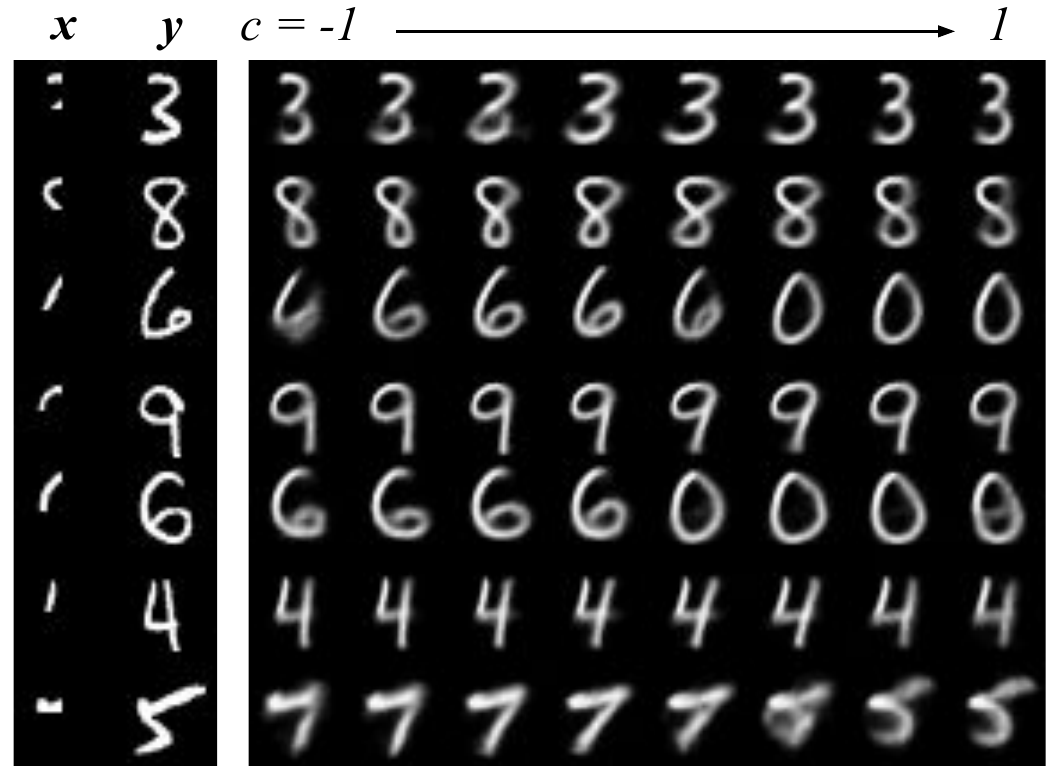}%
  \vspace{-5pt}
  \caption{\small Image completions using our algorithm with continuous values of \cc. 
  The first two columns show the input occluded \x values and the ground truth set of \Y values.
  On the right we show completions on this test image as \cc is swept from $-1$ to $1$. 
  }
  \label{fig:continuous_c}
  \vspace{-15pt}
\end{figure}

\paragraph{Continuous values for \cc}
Figure~\ref{fig:continuous_c} shows image completions from a continuous value of \cc.
Treating \cc as a continuous variable has the effect of ordering the completions, and allowing them to be smoothly interpolated.
This can, however, lead to implausible completions between modes, \eg in the final row where the $4$ merges into a $6$.

\begin{figure}
  \includegraphics[width=0.85\columnwidth,clip=true,trim={0.5cm 0.4cm 4.5cm 0.5cm}]{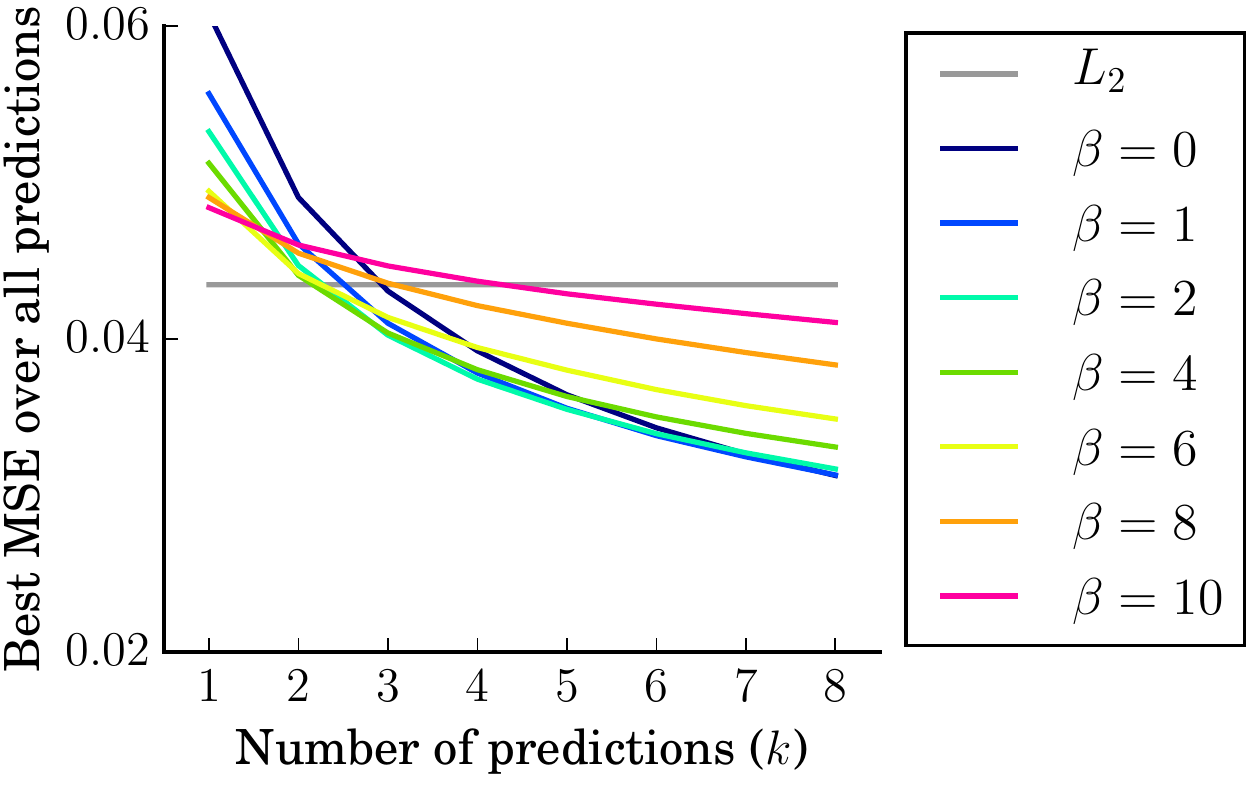}%
  \vspace{-5pt}
  \caption{\small The effect of varying $\beta$ on the MNIST dataset.
  We include the results for $L_2$ (in gray) for comparison with Figure \ref{fig:mnist_quant}. In general, we can see that as $\beta$ increases, the line tends towards the $L_2$ result; \ie error is improved for fewer samples, at the expense of reduced diversity.}
  \vspace{-18pt}
  \label{fig:beta_introspection}
\end{figure}

\paragraph{The effect of adjusting $\beta$}
\label{sec:beta_eval}
All results shown in this paper use $\beta=1$ (Section \ref{sec:beta}).
%However, we have experimented with adjusting this parameter on the MNIST completion task.
% A graph of this experiment is given in the supplementary material
Figure \ref{fig:beta_introspection} shows the quantitative effect of adjusting this parameter.
As $\beta$ increases, the curves tend toward the $L_2$ result; \ie error is improved for fewer samples, at the expense of reduced diversity.
A user of our system may wish to choose $\beta$ to suit the task at hand.

\begin{figure}[b]
\centering
  \vspace{-2pt}
  \includegraphics[width=1.0\columnwidth]{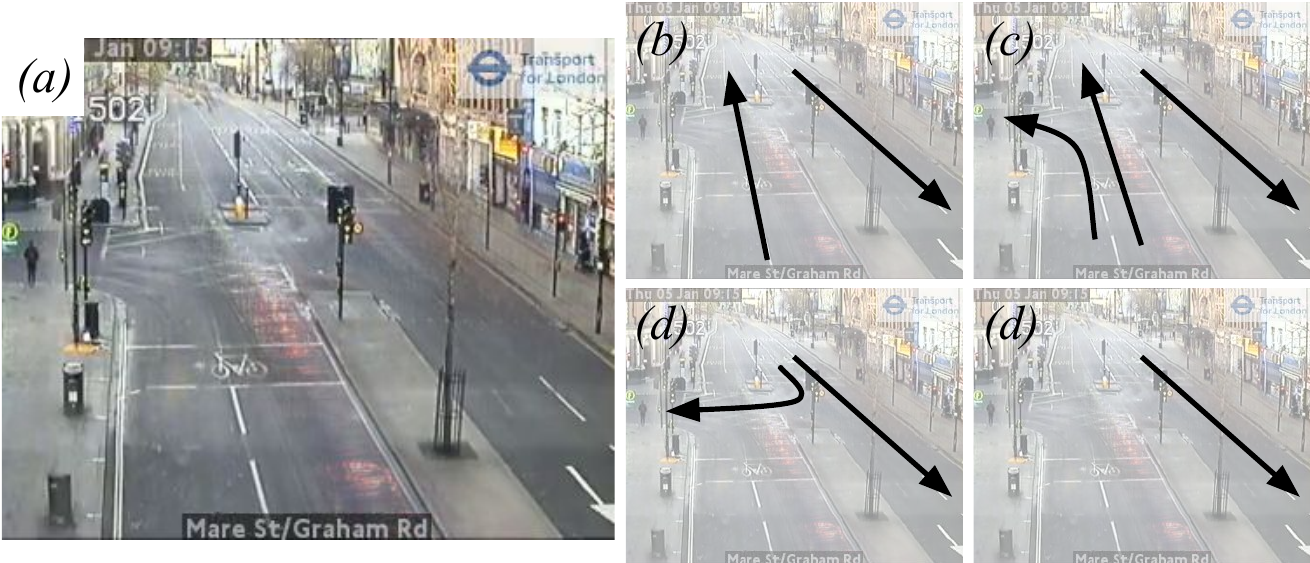}%
  \vspace{-5pt}
  \caption{\small Traffic flow dataset. Given an image of road junction (a), and some prior knowledge (\eg that this image is from a country with left-hand-traffic), one can envisage different patterns of traffic motion that might be expected. For example, (b) shows two-way traffic; (c) shows two-way traffic plus a left filter, etc.}
   \vspace{-5pt}
  \label{fig:tfl_problem_setup}
\end{figure}

%%%%%%%%%%%%%%%%%%%%%%%%%%%%%%%%%%%%%%%%%%%%%%%%%%%%%%%%%%%%
\subsection{Predicting traffic flows}

\begin{figure*}
\centering
  \includegraphics[width=2.1\columnwidth]{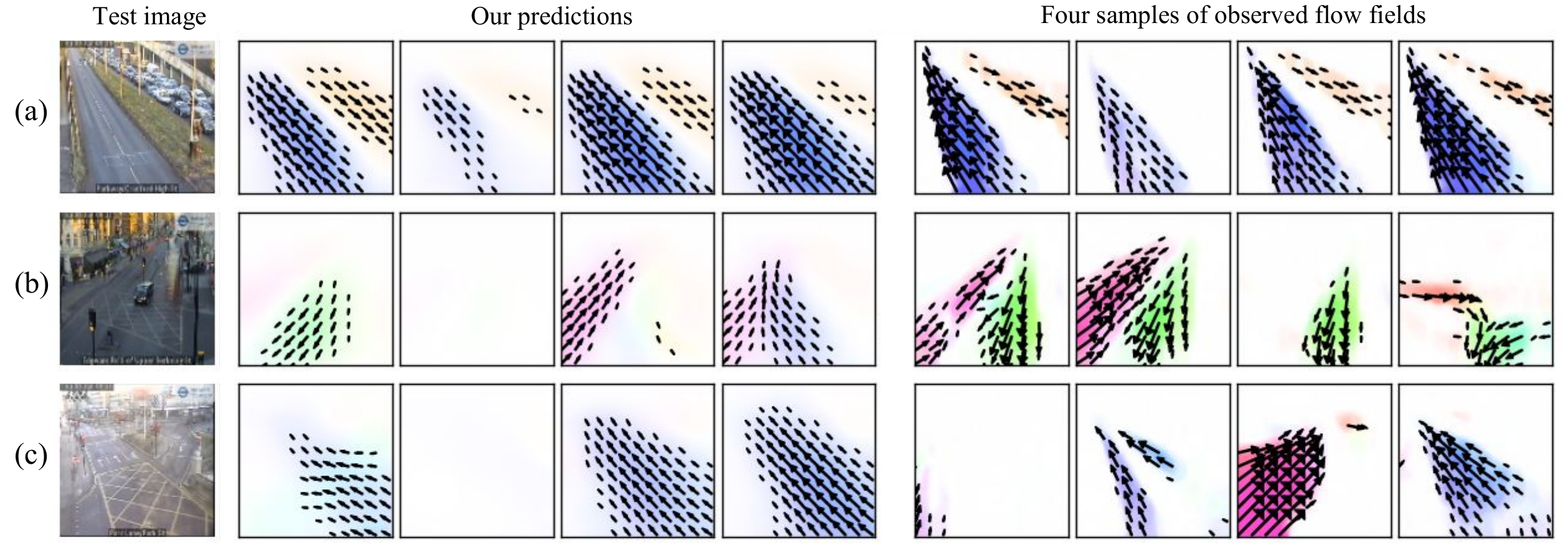}%
  \vspace{-8pt}
  \caption{\small Predictions on traffic flow dataset.  On the left is the input to our algorithm, a median image from a short traffic camera video. Test cameras are excluded from training. %unseen during training time.
  In the center are four flow predictions from our model, made only from the single test-time frame, while the right shows four of the many flow-fields actually observed by this traffic camera.
  In (a) our system predicts different volumes of traffic flowing in each direction of this two-way road.
  (b) shows a junction. As well as traffic flowing in each direction, we see traffic leaving (column 3) and joining (column 4) the main road. Here, column 2 predicts no traffic flowing, which is often the case in these short video clips.
  In (c), we correctly identify the road as a one-way street (cols.~3 and 4) with a separate road crossing the main flow (col.~1).
  }
  \vspace{-10pt}
  \label{fig:flow_res}
\end{figure*}

City traffic can exhibit diverse behaviors.
In this experiment, we tackle the problem: given a single image of a new road or intersection, can we predict what flow \emph{patterns} the traffic may exhibit?
Consider the image of the traffic intersection in Figure \ref{fig:tfl_problem_setup}.
Based on our knowledge, we can predict how traffic \emph{might} move, were it to be present.
There are many correct answers, depending on the phase of the traffic lights, traffic density, and the whims of individual drivers.

% \begin{figure}
%   \includegraphics[width=1.0\columnwidth,clip=true,trim={0.5cm 0.5cm 1cm 0.5cm}]{images/tfl_graph}%
%   \vspace{-5pt}
%   \caption{\small Flow prediction evaluation.  Here, we plot the best endpoint error over all the predictions made from each model, evaluated against each ground truth flow image.  We see that as more predictions are made, all methods improve; our full method outperforms all others.}
%   \label{fig:tfl_graph}
% \end{figure}

\begin{table}[b]
{
\small
\begin{tabular}{lcccc}
\toprule
\multicolumn{1}{r}{$k=$}  & 1 & 2 & 3 & 4 \\
\midrule
Lee et al. & 0.313 &  0.215 &  0.172 &  0.158 \\
$L_2$ + test-time Dropout & 0.209 &  0.194 &  0.187 &  0.184 \\
% Ours (no Dropout) & 0.232 &  0.195 &  0.170 &  0.151 & 
Ours & \textbf{0.203} &  \textbf{0.174} &  \textbf{0.158} &  \textbf{0.146} \\
  \bottomrule
  \\  \end{tabular}
  }
  \vspace{-5pt}
  \caption{\small Flow prediction evaluation. Lower numbers are better. Here, we report the best endpoint error over all the predictions made from each model, evaluated against each ground truth flow image.  We see that as more predictions are made, all methods improve, while our full method outperforms the baselines.}
  \vspace{-18pt}
  \label{tab:tfl_graph}
\end{table}

We created a dataset to learn about traffic flows, formed from short ({\raise.17ex\hbox{$\scriptstyle\sim$}}8s) videos taken from a publicly accessible traffic camera network \cite{tfl-jamcam}.
Each of the 912 traffic cameras in the network continuously films road traffic from a fixed viewpoint, and short clips of traffic are uploaded to their servers every five minutes. 
We obtained a total of 10,527 distinct videos, around 11.5 videos for each camera location.
For each video, we compute the average frame-to-frame optical flow (using \cite{weinzaepfel-iccv-2013}) as a single \y pattern.
We use a simple bottleneck architecture based on VGG16~\cite{simonyan-arxiv-2014} --- see supplemental materials for details.
We divide the camera locations into an 80/20 train/test split. To evaluate, we use the average endpoint error, as advocated by \cite{baker-ijcv-2011}. On this task, we find that each of \cite{lee-nips-2016}'s ensemble members produces a meaningful output, so we do not include $\epsilon L_2$. We report results compared to \cite{lee-nips-2016} and test-time dropout, as those were the best competitors from Section~\ref{sec:mnist_eval}.
Quantitative comparisons are given in Table~\ref{tab:tfl_graph}. We outperform the baselines, and representative results are depicted in Figure~\ref{fig:flow_res}. 
Note that while infinite combinations of vehicle-motion are possible (within physical limits), our predictions seek to capture the diverse modes, and to associate them with the appearance of regions in a single image.
Our diverse predictions capture (a) different directions in the major flow of traffic, (b) different densities of traffic flow, and (c) traffic flow into and from side roads.

%%%%%%%%%%%%%%%%%%%%%%%%%%%%%%%%%%%%%%%%%%%%%%
\subsection{ShapeNet volumes from silhouettes}
\label{sec:shapenet}

3D volume prediction from 2D silhouettes has recently become a popular task \cite{marrnet, ravi2017projectionnet}.
However, it is an inherently ambiguous problem, as many different volumes could produce any given silhouette image.
This task calls for diverse predictions: we expect that a diverse prediction network will be able to produce several different plausible 3D shape predictions for a given silhouette input.

\begin{figure*}
\centering
  \includegraphics[width=2.0\columnwidth]{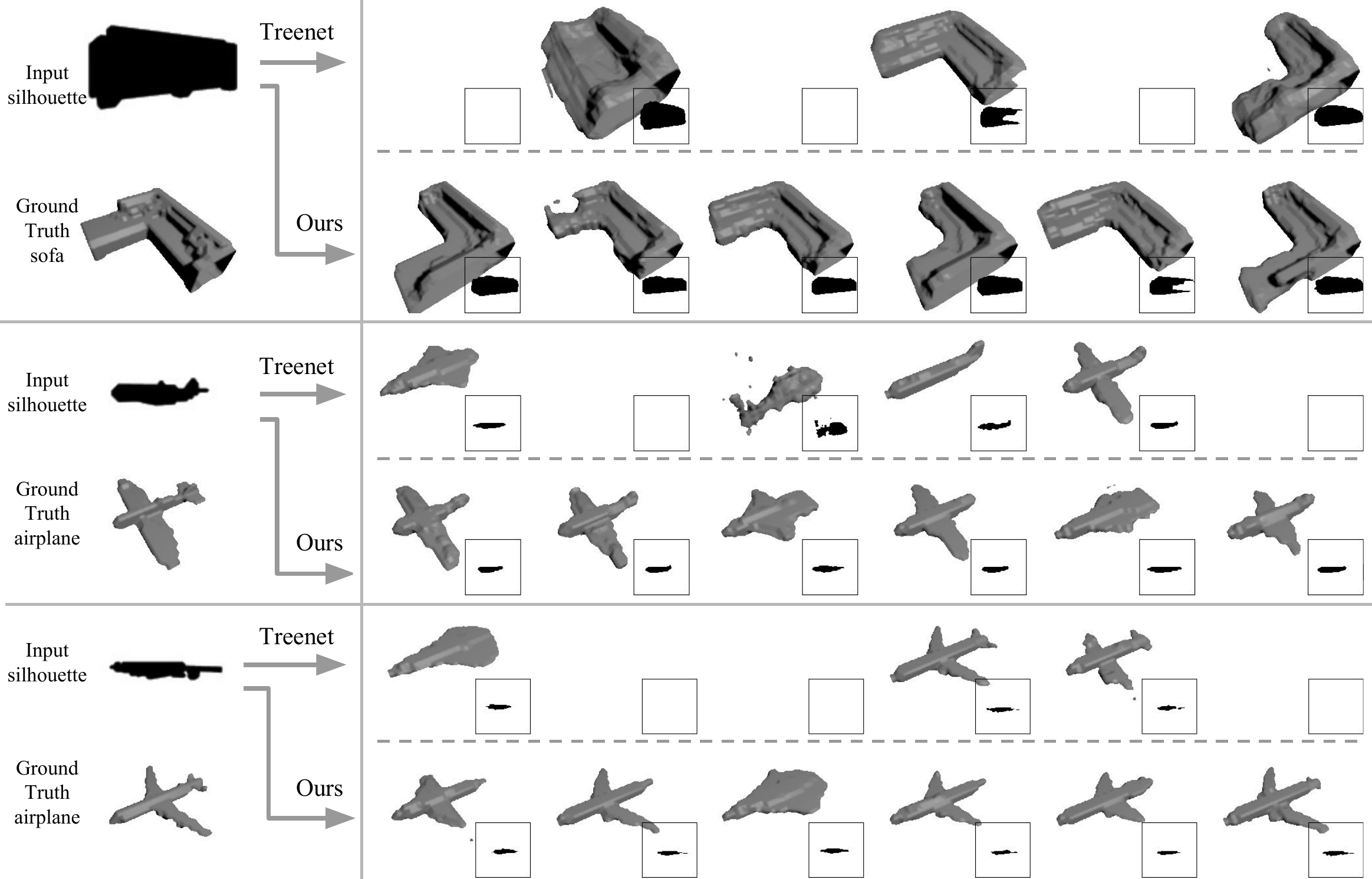} \\
  \vspace{-2pt}
  \caption{\small Predictions on Shapenet, comparing our algorithm with the Treenet architecture \cite{lee-arxiv-2015}.
  Many of the Treenet predictions are blank, as the network has failed to make a prediction from this ensemble member, while our network balances plausibility and variety to make a prediction for each value of \cc.
  Inset boxes for each prediction show that shape re-projected as a silhouette from the input camera's viewpoint. 
  More results are given in the supplementary material.
  }
  \vspace{-9pt}
  \label{fig:shapenet_qual}
\end{figure*}

We used a subset of three classes from ShapeNet \cite{3DShapeNets}, \texttt{airplane}, \texttt{car} and \texttt{sofa}, and we trained a system to predict the $32^{3}$ voxel grid given a single 2D silhouette rendering as input.
Training and testing silhouettes were rendered with a fixed elevation angle ($0^{\circ}$) with azimuth varying in $15^{\circ}$ increments.
We trained our model and baselines to produce 6 outputs at test time.
Details of architectures are given in the supplemental material.
Results are summarized in Table~\ref{tab:shapenet_results} and a few samples illustrated in Figure~\ref{fig:shapenet_qual}. 
Our results are typically better, and are achieved with far fewer parameters.
%It is worth noting that our DiverseNet architecture modification and the specialized loss aggregate to give numerically and qualitatively better results than the strongest baseline we could devise.
% we use the splits from: \cite{yan-arxiv-2017}

\paragraph{Examining the effect of architecture and loss}
We performed an ablation study to discover if the prevention of mode-dropping from our method comes about due to our architecture, or our $L_{catchup}$ loss.
We directly compared the ShapeNet results when training with our architecture vs Treenet, and for each architecture we turned $L_{catchup}$ on and off.
We also tried pretraining the networks (with the standard loss), to help each ensemble member to produce plausible outputs.
We find that, numerically, our proposed method (indicated with $\Rightarrow$) outperforms all other variants.

\paragraph{Measuring plausibility of predictions}
%For each test image, there is just one ground truth voxel volume, while our network can produce multiple predictions.
In this task, one method of assessing overall plausibility of the predictions is to measure their compatibility with the input silhouettes.
The intuition is that good predictions will match with the input silhouette, even if they don't necessarily match the ground truth.
We convert each predicted grid to a mesh \cite{lorensen-siggraph-1987} and render using the same camera parameters used to create the original silhouette.
This is compared to the input silhouette using the IoU, and the average over all of these IoU measures is given in Table \ref{tab:shapenet_results}.
The Treenet architecture tends to give a lower re-projection IoU than equivalent predictions from our architecture.
This is because of mode dropping; predictions from their network are not always plausible.

\newcommand{\bb}[1]{\textbf{#1}}
\newcommand{\bbc}[1]{\textcolor{darkred}{\textbf{#1}}}
\newcommand{\col}[1]{\textcolor{darkred}{#1}}

\begin{table*}
\centering
\footnotesize
\begin{tabular}{cccc|cc|cccccc|cc}
\toprule
\bb{Architecture}
& \bb{Params}
& \bb{Test-time}
& & $L_{catchup}$ 
& \bb{Pretrain}
& \textbf{k=1} & \textbf{k=2} & \textbf{k=3} & \textbf{k=4} & \textbf{k=5} & \textbf{k=6} & \textbf{Reproj} & \textbf{Var} \\ 
 \midrule 
\multirow{4}{*}{Treenet \cite{lee-nips-2016}}
& \multirow{4}{*}{66.8M}
& \multirow{4}{*}{0.047s}
& & &
& 0.296 & 0.376 & 0.435 & 0.482 & 0.518 & 0.545 & 0.313  & 0.060  \\ 
& & & & $\bullet$ &
& 0.650 & 0.665 & 0.676 & 0.683 & 0.689 & 0.694 & 0.726  & 0.015  \\ 
& & & &  & $\bullet$
& 0.363 & 0.454 & 0.516 & 0.558 & 0.586 & 0.607 & 0.502  & 0.059  \\ 
& & & & $\bullet$ & $\bullet$
& 0.666 & 0.679 & 0.688 & 0.694 & 0.700 & 0.704 & 0.733  & 0.013  \\ 
\midrule 
\multirow{4}{*}{DiverseNet}  
& \multirow{4}{*}{12.6M}
& \multirow{4}{*}{0.053s}
& & &
& 0.661 & 0.680 & 0.692 & 0.701 & 0.708 & 0.713 & \bb{0.773}  & 0.014  \\ 
& & & $\Rightarrow$
& \cellcolor{lightred} $\bullet$  
& \cellcolor{lightred}
& \cellcolor{lightred} \bb{0.680} 
& \cellcolor{lightred} \bb{0.693} 
& \cellcolor{lightred} \bb{0.702} 
& \cellcolor{lightred} \bb{0.708} 
& \cellcolor{lightred} \bb{0.713} 
& \cellcolor{lightred} \bb{0.716} 
& \cellcolor{lightred} 0.753  
& \cellcolor{lightred} 0.010 \\ 
& & & & & $\bullet$
& 0.650 & 0.669 & 0.682 & 0.692 & 0.698 & 0.704 & 0.754  & 0.016  \\ 
& & & & $\bullet$ & $\bullet$
& 0.671 & 0.678 & 0.683 & 0.687 & 0.690 & 0.693 & 0.757  & 0.003  \\ 
\bottomrule 
\end{tabular}
\vspace{-5pt}
\caption{Quantitative results on Shapenet volume prediction, where all numbers are averaged across three classes.
The numbers for $k=1, \ldots 6$ are the IoU on the predicted voxel grids.
'Reproj' is the IoU computed on the silhouette reprojection (see text for more details).
The row marked with $\Rightarrow$ and {\setlength{\fboxsep}{0pt}\colorbox{lightred}{highlighted red}} is our full method as described in this paper.
All other rows are baselines and ablations, and show that both our loss and our architecture combine to achieve improved results, and that we do not need pretraining.
Variance is listed for completeness, but it is a false measure of ``diversity,'' crediting even unrepresentative $3$D shapes.
}
\label{tab:shapenet_results}
%\vspace{-5pt}
\end{table*}

\newcommand{\ruler}{\textcolor[RGB]{180,180,180}{\noindent\rule{\linewidth}{0.2pt}} \\}

\section{Conclusions} 

We have presented a novel way of training a single network to map a single test-input to multiple predictions. 
This approach has worked for predicting diverse appearance, motion, and 3D shape. 
We are not the first to explicitly propose diversity modeling, but a key advantage of our approach is its simplicity, which yields good scores with small models.
%It can be applied to any supervised learning architecture.
Through a minimal modification to the loss function and training procedure, applicable to all network architectures, we can readily upgrade existing models to produce state-of-the-art diverse outputs without the need for expensive sampling or ensemble approaches.

%We obtain diversity at the expense of a small drop in raw accuracy when measured on a single prediction.

%In addition, our method allows the network to exploit training-time diversity to learn the range of predictions that should be made. 
The impact of our innovation is greatest when ground-truth labels are multi-modal, so where two big modes are averaged by the model, or minority modes are overlooked. 
We anticipate that our method will have applications in user-in-the-loop scenarios, where predictions can be presented as multiple-choice options to a user \cite{marks-acm-1997}.
%This is an important consideration  in computer vision, especially considering biases in datasets.

%We have evaluated on four challenging tasks, and there are many other ambiguous tasks that stand to benefit, such as 3D volume completion \cite{song-cvpr-2017} and learning of rendering functions \cite{nalbach-arxiv-2016}.
%

\paragraph{Limitations}
Like unsupervised methods such as $k$-means \cite{steinhaus-bpas-1957}, our number of predictions must be user-specified at training time. 
We leave determining an optimum value for this parameter (similar to \cite{pelleg-icml-2000}) as future work.

\vspace{6pt}

\paragraph{Acknowledgements}
This work has been supported by the SecondHands project,
funded from the EU Horizon 2020 Research and Innovation
programme under grant agreement No 643950, NERC NE/P019013/1, Fight for Sight UK, and EPSRC CAMERA (EP/M023281/1).

\clearpage

{\fontsize{9}{10}\selectfont
\bibliographystyle{ieee}
\bibliography{bibtex/strings.bib,bibtex/main.bib}
}

\end{document}